\documentclass[11pt,a4paper]{article}
\usepackage[hyperref]{acl2020}
\usepackage{times}
\usepackage{latexsym}

\usepackage{microtype}

\usepackage{booktabs,colortbl}
\usepackage{graphicx}
\usepackage{subcaption}
\usepackage{multirow}
\usepackage{todonotes}
\usepackage{wrapfig}

\usepackage{url}
\usepackage{amsmath,amssymb,physics,xspace}
\usepackage{verbatim}
\usepackage{array}

\usepackage{color}

\newcolumntype{H}{>{\setbox0=\hbox\bgroup}c<{\egroup}@{}}

\newcommand{\x}{\ensuremath{x}}

\newcommand{\xh}{\ensuremath{x_{<i}}}

\newcommand{\mat}[1]{\ensuremath{\mathbf #1}}

\DeclareMathOperator*{\dims}{\textit{d}}

\DeclareMathOperator*{\Cat}{Cat}

\DeclareMathOperator*{\KL}{KL}

\DeclareMathOperator*{\JS}{JS}

\DeclareMathOperator*{\ELBO}{ELBO}
\DeclareMathOperator*{\diag}{diag}
\DeclareMathOperator*{\softmax}{softmax}

\DeclareMathOperator*{\softplus}{softplus}
\DeclareMathOperator*{\affine}{affine}
\DeclareMathOperator*{\BiGRU}{BiGRU}

\DeclareMathOperator*{\GRU}{GRU}
\DeclareMathOperator*{\emb}{emb}
\DeclareMathOperator*{\layer}{layer}

\DeclareMathOperator{\R}{\mathbb R}

\aclfinalcopy %
\def\aclpaperid{3338} %

\title{Effective Estimation of Deep Generative Language Models}

\author{Tom Pelsmaeker\thanks{Work done while the first author was at the University of Amsterdam. Code is available at \url{https://github.com/tom-pelsmaeker/deep-generative-lm}} \\
  ILCC \\
  University of Edinburgh \\
  \texttt{t.l.pelsmaeker@sms.ed.ac.uk} \\\And
  Wilker Aziz \\
  ILLC\\
  University of Amsterdam\\
  \texttt{w.aziz@uva.nl} \\}

\date{}

\renewcommand\footnotemark{}

\begin{document}
\maketitle
\begin{abstract}
    Advances in variational inference enable parameterisation of probabilistic models by deep neural networks. This combines the statistical transparency of the probabilistic modelling framework with the representational power of deep learning. 
    Yet, due to a problem known as \emph{posterior collapse}, it is difficult to estimate such models in the context of language modelling effectively. %
    We concentrate on one such model, the variational auto-encoder, which we argue is an important building block in hierarchical probabilistic models of language. 
    This paper contributes a sober view of the problem, a survey of techniques to address it, novel techniques, and extensions to the model. 
    To establish a ranking of techniques, we perform a systematic comparison using Bayesian optimisation and find that many techniques perform reasonably similar, given enough resources. Still, a favourite can be named based on convenience. We also make several empirical observations and recommendations of best practices that should help researchers interested in this exciting field.
\end{abstract}

\section{Introduction}
Deep generative models (DGMs) are probabilistic latent variable models parameterised by neural networks (NNs). Specifically, DGMs optimised with amortised variational inference and reparameterised gradient estimates \citep{Kingma+2014:VAE,RezendeEtAl14VAE}, better known as variational auto-encoders (VAEs), have spurred much interest in various domains, including computer vision and natural language processing (NLP).

In NLP, VAEs have been developed for word representation \citep{rios18}, morphological analysis \citep{morphanalysis}, syntactic  and semantic  parsing \citep{corro2019,lyu-titov:2018:Long}, document modelling \citep{miao2016neural}, summarisation \citep{MiaoSummary},  machine translation \citep{VNMT,SDEC,eikema-aziz-2019-auto},  %
 language and vision \citep{Yunchenetal2016_NIPS,Wangetal2017_NIPS},  %
 dialogue modelling \citep{wen2017latent,serban2017hierarchical}, 
 speech modelling \citep{fraccaro2016sequential}, 
 and, of course, language modelling \citep{bowman2016generating,goyal2017z}.
One problem remains common to the majority of these models, VAEs often learn to ignore the latent variables.

We investigate this problem, dubbed  \emph{posterior collapse}, %
in the context of language models (LMs). 
In a deep generative LM \citep{bowman2016generating}, sentences are generated conditioned on samples from a continuous latent space, an idea with various practical applications. %
For example, one can constrain %
this latent space to promote generalisations that are in line with linguistic knowledge and intuition \citep{xu2018spherical}. 
This also allows for greater flexibility in how the model is used, for example, to generate sentences that live---in latent space---in a neighbourhood of a given observation \citep{bowman2016generating}. %
Despite this potential, VAEs that employ strong generators (e.g. recurrent NNs) tend to ignore the latent variable. %
Figure \ref{fig:vanilla} illustrates this point: %
neighbourhood in latent space does not correlate to patterns in data space, and the model %
behaves just like a standard LM.

Recently, many techniques have been proposed to address this problem (\S\ref{sec:collapse} and \S\ref{sec:related}) and they range from modifications to the objective to changes to the actual model. Some of these techniques have only been tested under different conditions and under different evaluation criteria, and some of them have only been tested outside NLP.
This paper contributes: %
(1) a novel strategy based on constrained optimisation towards a pre-specified upper-bound on mutual information; %
(2) multimodal priors that by design promote increased mutual information between data and latent code; last and, arguably most importantly, (3) a systematic comparison---in terms of resources dedicated to hyperparameter search and sensitivity to initial conditions---of strategies to counter posterior collapse, including some never tested for language models (e.g. InfoVAE, LagVAE, soft free-bits, and  multimodal priors).

\begin{figure*}[t]
\centering
\begin{subfigure}[t]{0.46\textwidth}
\centering
\small
\begingroup
\setlength{\tabcolsep}{4pt} %
\begin{tabular}{@{}l p{5.8cm}@{}}
\toprule 
Decoding & Generated sentence \\ \midrule
Greedy &  The company said it expects to report net income of \$UNK-NUM million\\ \midrule
Sample & They are getting out of my own things ?  \\
& IBM also said it will expect to take next year .  \\ \bottomrule
\end{tabular}
\endgroup
\caption{\label{fig:vanilla-samples}Greedy generation from prior samples (top) yields the same sentence every time, showing that the latent code is ignored. Yet, ancestral sampling (bottom) produces good sentences, showing that the recurrent decoder learns about the structure of English sentences.} %
\end{subfigure}
~
\begin{subfigure}[t]{0.52\textwidth}
\centering
\small
\begin{tabular}{@{}p{\textwidth}@{}}
\toprule
\textbf{The two sides hadn't met since Oct. 18.} \\
I don't know how much money will be involved. \\
The specific reason for gold is too painful. \\
The New Jersey Stock Exchange Composite Index gained 1 to 16. \\
And some of these concerns aren't known. \\
\textbf{Prices of high-yield corporate securities ended unchanged.} \\ \bottomrule 
\end{tabular}
\caption{Homotopy: ancestral samples mapped from points along a linear interpolation of two given sentences as represented in latent space. The sentences do not seem to exhibit any coherent relation, showing that the model does not exploit neighbourhood in latent space to capture regularities in data space.} %
\label{fig:vanilla-hom}
\end{subfigure}
\caption{\label{fig:vanilla}Sentences generated from \citet{bowman2016generating}'s VAE trained \emph{without} special treatment.}
\end{figure*}

\section{Density Estimation for Text}

Density estimation for written text has a long history \citep{jelinek1980interpolated,Goodman:2001:BPL},  %
but in this work we concentrate on neural network models \citep{bengio2003neural}, %
 in particular, autoregressive ones \citep{mikolov2010recurrent}.
Following common practice, we model sentences independently, each a sequence $x = \langle x_1, \ldots, x_n \rangle$ of $n = \abs{x}$ tokens. %

\subsection{Language models}

A language model (LM) prescribes the generation of a sentence as a sequence of categorical draws parameterised in context, i.e. $P(x|\theta)=$ %
\begin{equation}
\begin{aligned}
\prod_{i=1}^{\abs{x}} P(x_i|\xh, \theta) = \prod_{i=1}^{\abs{x}} \Cat(x_i|f(\xh; \theta)) ~ .
\end{aligned}
\end{equation}
To condition on all of the available context, a fixed NN $f(\cdot)$  %
maps from a prefix sequence (denoted $\xh$) to the parameters of a categorical distribution over the vocabulary. %
We  estimate the parameters $\theta$ of the model by searching for a local optimum of the log-likelihood function $\mathcal L(\theta) \! = \!\mathbb E_{X}[\log P(\x|\theta)]$ via stochastic gradient-based optimisation \citep{robbins1951stochastic,BottouCun2004}, where the expectation is taken w.r.t. the true data distribution and approximated with samples $x \sim \mathcal D$ from a data set of i.i.d. observations.
Throughout, we refer to this model as \textsc{RnnLM} alluding to a particular choice of $f(\cdot;\phi)$ that employs a recurrent neural network \citep{mikolov2010recurrent}. %

\subsection{Deep generative language models}

\citet{bowman2016generating} model observations as draws from the marginal of a DGM. 
An NN maps from a latent sentence embedding $z \in \mathbb R^{\dims_z}$ to a distribution $P(x|z, \theta)$ over sentences,
\begin{equation}
\begin{aligned}
    &P(\x|\theta) = \int p(z)P(\x|z, \theta) \dd z \\
    &= \int \!\mathcal N(z|0, I)\prod_{i=1}^{\abs{\x}} \Cat(x_i|f(z, \xh; \theta)) \dd z ~,
\end{aligned}
\end{equation}
where $z$ follows a standard Gaussian prior.\footnote{We use uppercase $P(\cdot)$ for probability mass functions and lowercase $p(\cdot)$ for probability density functions.} 
Generation still happens one word at a time without Markov assumptions, but $f(\cdot)$ now conditions on $z$ in addition to the observed prefix. 
The conditional $P(x|z, \theta)$ is commonly referred to as \emph{generator} or \emph{decoder}. The quantity $P(x|\theta)$ is the \emph{marginal likelihood}, essential for parameter estimation.

This model is trained to assign a high (marginal) probability to observations, much like standard LMs. 
Unlike standard LMs, it employs a latent space which can accommodate a low-dimensional manifold where discrete sentences are mapped to, via posterior inference $p(z|x, \theta)$, and from, via generation $P(x|z, \theta)$. 
This gives the model an explicit mechanism to exploit neighbourhood and smoothness in latent space to capture regularities in data space. For example, it may group sentences according to latent factors (e.g. lexical choices, syntactic complexity, etc.).
It also gives users a mechanism to steer generation towards a specific purpose. For example, one may be interested in generating sentences that are mapped from the neighbourhood of another in latent space. %
To the extent this embedding space captures appreciable regularities, interest in this property is heightened. 

\paragraph{Approximate inference} Marginal inference for this model is intractable and calls for  variational inference  \citep[VI;][]{Jordan+1999:VI}, whereby an auxiliary and independently parameterised model $q(z|\x, \lambda)$ approximates the true posterior $p(z|\x, \theta)$. 
When this \emph{inference model} is itself parameterised by a neural network, we have a case of \emph{amortised inference} \citep{Kingma+2014:VAE,RezendeEtAl14VAE} and an instance of what is known as a VAE.
\citet{bowman2016generating} approach posterior inference with a  Gaussian model
\begin{equation}
\begin{aligned}
    Z|\lambda, \x &\sim \mathcal N(\mathbf u, \diag(\mathbf s \odot \mathbf s))\\
    [\mathbf u, \mathbf s] &= g(\x; \lambda)
\end{aligned}
\end{equation}
whose parameters, i.e. a location vector $\mathbf u \in \mathbb R^D$ and a scale vector $\mathbf s \in \mathbb R^D_{>0}$, are predicted by a neural network architecture $g(\cdot; \lambda)$ from an encoding of the complete observation $\x$.\footnote{We use boldface for deterministic vectors and $\odot$ for elementwise multiplication.}
In this work, we use a bidirectional recurrent encoder.  %
Throughout the text we will refer to this model as \textsc{SenVAE}.

\paragraph{Parameter estimation} We can jointly estimate the parameters of both models (i.e. generative and inference) by locally maximising a lower-bound on the log-likelihood function (ELBO)
\begin{equation}
\begin{aligned}
\mathcal E(\theta, \lambda) = \mathbb E_X\big[ \mathbb E_{q(z|\x, \lambda)}\left[ \log P(\x|z, \theta) \right]  \\
 - \KL(q(z|\x, \lambda)||p(z)) \big] ~ .
\end{aligned}
\end{equation}
For as long as we can reparameterise samples from $q(z|x,\lambda)$ using a fixed random source, automatic differentiation \citep{BaydinEtAl2015AD} can be used to obtain unbiased gradient estimates of the ELBO \citep{Kingma+2014:VAE,RezendeEtAl14VAE}. %

\section{\label{sec:collapse}Posterior Collapse}

In VI, we make inferences using an approximation $q(z|x, \lambda)$ to the true posterior $p(z|x, \theta)$ and choose $\lambda$ as to minimise the KL divergence $\mathbb E_X[\KL(q(z|x, \lambda)||p(z|x, \theta))]$. 
The same principle yields a lower-bound on log-likelihood used to estimate $\theta$ jointly with $\lambda$, thus making the true posterior $p(z|x, \theta)$ a moving target. 
If the estimated conditional $P(x|z, \theta)$ can be made independent of $z$, which in our case means relying exclusively on $x_{<i}$ to predict the distribution of $X_i$, the true posterior will be independent of the data and equal to the prior.\footnote{This follows trivially from the definition of posterior: $p(z|x) = \frac{p(z)P(x|z)}{P(x)} \overset{X \perp Z}{=} \frac{p(z)P(x)}{P(x)} = p(z)$.}
Based on such observation, \citet{chen2016variational} argue that information that can be modelled by the generator without using latent variables will be modelled that way---precisely because when no information is encoded in the latent variable the true posterior equals the prior and it is then trivial to reduce $\mathbb E_X[\KL(q(z|x, \lambda)||p(z|x, \theta))]$ to $0$.
This is typically diagnosed by noting that after training $\mathbb \KL(q(z|\x, \lambda)||p(z))\! \to\! 0$ for most $x$: we say that \emph{the true posterior collapses to the prior}. 
\citet{alemi2018fixing} show that the \emph{rate}, $R = \mathbb E_X[\KL(q(z|x,\lambda)||p(z))]$, 
is an upperbound to $I(X;Z|\lambda)$, the mutual information (MI) between $X$ and $Z$. Thus, if $\KL(q(z|x,\lambda)||p(z))$ is close to zero for most training instances, MI is either $0$ or negligible. 
They also show that the \emph{distortion}, $D = -\mathbb E_X[ \mathbb E_{q(z|x,\lambda)}[\log P(x|z,\theta)]]$, relates to a lower-bound on MI (the lower-bound being $H - D$, where $H$ is the unknown data entropy).

A generator that makes no Markov assumptions, such as a recurrent LM, can potentially achieve $X_i \perp Z \mid x_{<i}, \theta$, and indeed many have noticed that VAEs whose observation models are parameterised by such \emph{strong generators} (or strong decoders) tend to ignore the latent representation \citep{bowman2016generating,higgins2016beta,sonderby2016ladder,zhao2017infovae}. %
For this reason, a strategy to prevent posterior collapse is to weaken the decoder \citep{yang2017improved,semeniuta2017hybrid, park2018dialogue}. %
In this work, we are interested in employing strong generators, thus we do not investigate weaker decoders.
Other strategies involve changes to the optimisation procedure and manipulations to the objective that target local optima of the ELBO with non-negligible MI.

\paragraph{Annealing} \citet{bowman2016generating} propose ``KL annealing'', whereby the $\KL$ term in the $\ELBO$ is incorporated into the objective in gradual steps. 
This way the optimiser can focus on reducing distortion early on in training, potentially by increasing MI.
They also propose to drop words from $x_{<i}$ at random to weaken the decoder---intuitively the model would have to rely on $z$ to compensate for missing history.
We experiment with a slight modification of word dropout whereby we slowly vary the dropout rate from $1 \to 0$.  In a sense, we ``anneal'' from a weak to a strong generator.

\paragraph{Targeting rates} Another idea is to target a pre-specified rate \citep{alemi2018fixing}. %
\citet{kingma2016improved} replace the $\KL$ term in the ELBO with $\max(r, \KL(q(z|x, \lambda)||p(z)))$, dubbed \emph{free bits} (FB) because it allows encoding the first $r$ nats of information ``for free''. 
As long as $\KL(q(z|x, \lambda)||p(z)) < r$, this does not optimise a proper ELBO (it misses the $\KL$ term), and the $\max$ introduces a discontinuity. %
\citet{chen2016variational} propose \emph{soft free bits} (SFB), that instead multiplies the $\KL$ term in the ELBO with a weighing factor $0 < \beta \leq 1$ that is dynamically adjusted based on the target rate $r$: $\beta$ is incremented (or reduced) by $\omega$ if $R > \gamma r$ (or $R < \varepsilon r$). %
Note that this technique requires hyperparameters (i.e. $\gamma, \varepsilon, \omega$) besides $r$ to be tuned in order to determine how $\beta$ is updated.

\paragraph{Change of objective} We may also seek alternatives to the $\ELBO$ as an objective %
and relate them to quantities of interest such as MI.
A simple adaptation of the ELBO weighs its KL-term by a constant factor \citep[$\beta$-VAE;][]{higgins2016beta}. 
Setting $\beta < 1$ promotes increased MI. Whilst being a useful counter to posterior collapse, low $\beta$ might lead to variational posteriors becoming point estimates. 
InfoVAE \citep{zhao2017infovae} mitigates this with a term aimed at minimising the divergence from the \textit{aggregated} posterior $q(z|\lambda)=\mathbb E_X[q(z|x,\lambda)]$ to the prior. %
Following \citet{zhao2017infovae}, we approximate this with an estimate of maximum mean discrepancy \citep[MMD;][]{gretton2012kernel} in our experiments.
Lagrangian VAE \citep[LagVAE;][]{zhao2018information} casts VAE optimisation as a dual problem; it targets either maximisation or minimisation of (bounds on) $I(X;Z|\lambda)$ under constraints on the InfoVAE objective. 
In MI-maximisation mode, LagVAE maximises a weighted lower-bound on MI, $-\alpha D$, under two constraints, a maximum -\textsc{ELBO} and a maximum \textsc{MMD}, that prevent $p(z|x,\theta)$ from degenerating to a point mass. Reasonable values for these constraints have to be found empirically.

\section{\label{sec:MDR}Minimum Desired Rate}

We propose \emph{minimum desired rate} (MDR), a technique to attain ELBO values at a pre-specified rate $r$ that does not suffer from the gradient discontinuities of FB, and does not introduce the additional hyperparameters of SFB. %
The idea is to optimise the ELBO subject to a minimum rate constraint $r$:
\begin{equation}
\begin{aligned}
    &\max_{\theta, \lambda} ~ \mathcal E(\theta, \lambda), \\
    &~\text{s.t. } \mathbb E_X\left[ \KL(q(z|x,\lambda)||p(z)) \right] > r ~ .
\end{aligned}
\end{equation}
Because constrained optimisation is generally intractable, we optimise the Lagrangian \citep{boyd2004convex} $\Phi(\theta, \lambda, u)=$ %
\begin{equation}
    \mathcal E(\theta, \lambda) - u(r - \mathbb E_X[\KL(q(z|x,\lambda)||p(z))])
\end{equation}
where $u \in \mathbb R_{\geq 0}$ is a positive Lagrangian multiplier.
We define the dual function $\phi(u) = \max_{\theta, \lambda} ~ \Phi(\theta, \lambda, u)$
and solve the dual problem $\min_{u \in \mathbb R_{\geq 0}} ~ \phi(u) $.
Local minima of the resulting min-max objective can be found by performing stochastic gradient descent with respect to $u$ and stochastic gradient ascent with respect to $\theta, \lambda$. %

\subsection{Relation to other techniques}

It is insightful to compare MDR to the various techniques we surveyed in terms of the gradients involved in their optimisation.
The losses minimised by $\KL$ annealing, $\beta$-VAE, and SFB have the form
$\ell_{\beta}(\theta, \lambda) = D + \beta R$,
where $\beta \ge 0$.
FB minimises the loss 
$\ell_{\text{FB}}(\theta, \lambda) = D + \max(r, R)$,
where $r > 0$ is the target rate. Last, with respect to $\theta$ and $\lambda$, MDR minimises the loss $\ell_{\text{MDR}}(\theta, \lambda) = D + R + u (r - R)$,
where $u \in \mathbb R_{\geq 0}$ is the Lagrangian multiplier. And with respect to $u$, MDR minimises 
$\phi(u) = - D - R - u (R - r)$.

\begin{subequations}
Let us inspect gradients with respect to the parameters of the VAE, namely, $\theta$ and $\lambda$.
FB's gradient %
$\grad_{\theta, \lambda} \ell_{\text{FB}}(\theta, \lambda) = $
\begin{equation}
    \grad_{\theta, \lambda} D + \begin{cases}
    0 & \text{if } R \leq r \\
    \grad_{\theta, \lambda} R & \text{otherwise}
    \end{cases} 
\end{equation}
is discontinuous, that is, there is a sudden `jump' from zero to a (possibly) large gradient w.r.t. $R$ when the rate dips above $r$. 
KL annealing, $\beta$-VAE, and SFB do not present such discontinuity
\begin{equation}
    \grad_{\theta, \lambda} \ell_{\beta}(\theta, \lambda) =\grad_{\theta, \lambda} D + \beta \grad_{\theta, \lambda} R ~,
\end{equation}
for $\beta$ scales the gradient w.r.t. $R$. %
The gradient of the MDR objective is
\begin{equation}
    \grad_{\theta, \lambda} \ell_{\text{MDR}}(\theta, \lambda) =  \grad_{\theta, \lambda} D + (1 - u) \grad_{\theta, \lambda} R 
\end{equation}
which can be thought of as $\grad_{\theta, \lambda} \ell_{\beta}(\theta, \lambda)$ with $\beta$ dynamically set to $1 - u$ at every gradient step. 
\end{subequations}

\begin{subequations}
Hence, MDR is another form of KL weighing, albeit one that targets a specific rate.
Compared to $\beta$-VAE, MDR has the advantage that $\beta$ is not fixed but estimated to meet the requirements on rate. %
Compared to $\KL$-annealing, MDR dispenses with a fixed schedule for updating $\beta$, not only annealing schedules are fixed, they require multiple decisions (e.g. number of steps, linear or exponential increments) whose impact on the objective are not directly obvious.  
Most similar then, seems SFB. Like MDR, it flexibly updates $\beta$ by targeting a rate. However, differences between the two techniques become apparent when we observe how $\beta$ is updated. In case of SFB:
\begin{align}
    \beta^{(t+1)} = \beta^{(t)} + \begin{cases}
        \omega & \text{if } R > \gamma  r \\
        -\omega & \text{if } R < \varepsilon r
    \end{cases}
\end{align}
where $\omega$, $\gamma$ and $\varepsilon$ are hyperparameters. 
In case of MDR (not taking optimiser-specific dynamics into account):
\begin{equation}
    u^{(t+1)} = u^{(t)} - \rho \pdv{\phi(u)}{u} = u^{(t)} + \rho (R - r)
\end{equation}
where $\rho$ is a learning rate. 
From this, we conclude that MDR is akin to SFB, but MDR's update rule is a direct consequence of Lagrangian relaxation and thus dispenses with the additional hyperparameters in SFB's handcrafted update rule.\footnote{Note that if we set $\gamma = 1$, $\varepsilon = 1$, and $\omega = \rho (R - r)$ at every step of SFB, we recover MDR.}
\end{subequations}

\section{\label{sec:expressive}Expressive Priors}

Suppose we employ a multimodal prior $p(z|\theta)$, e.g. a mixture of Gaussians, and suppose we employ a unimodal posterior approximation, e.g. the typical diagonal Gaussian. 
This creates a mismatch between the prior and the posterior approximation families that makes it impossible for  $\KL(q(z|x,\lambda)||p(z|\theta))$ to be precisely $0$. 
For the aggregated posterior $q(z|\lambda)$  to match the prior, the inference model would have to---on average---cover all of the prior's modes. Since the inference network is deterministic, it can only do so as a function of the conditioning input $x$, thus increasing $I(X;Z|\lambda)$. Admittedly, this conditioning might still only capture shallow features of  $x$, and the generator may still choose to ignore the latent code, keeping $I(X;Z|\theta)$ low, but the potential seems to justify an attempt.
This view builds upon  \citet{alemi2018fixing}'s information-theoretic view which suggests that the prior regularises the inference model capping $I(X; Z|\lambda)$.
Thus, we modify \textsc{SenVAE} to employ a more complex, ideally multimodal, parametric prior $p(z|\theta)$  and fit its parameters.

\paragraph{MoG} %
Our first option is a uniform mixture of Gaussians (MoG), i.e. $p(z|\theta)=$
\begin{equation}
    \frac{1}{C} \sum_{c=1}^C \mathcal N(z|\boldsymbol\mu^{(c)}, \diag(\boldsymbol\sigma^{(c)} \odot \boldsymbol\sigma^{(c)}))
\end{equation}
where the Gaussian parameters are optimised along with other generative parameters. 
Note that though we give this prior up to $C$ modes, the optimiser might merge some of them (by learning approximately the same location and scale). 

\paragraph{VampPrior} Motivated by the fact that, for a fixed posterior approximation, the prior that optimises the $\ELBO$ equals $\mathbb E_X[q(z|\x, \lambda)]$,  \citet{tomczak2017vae} propose the VampPrior, 
a \emph{variational mixture of posteriors}: 
\begin{equation}
    p(z|\theta) = \frac{1}{C} \sum_{c=1}^C q(z|v^{(c)}, \lambda)
\end{equation}
where $v^{(c)}$ is a learned pseudo input---in their case a continuous  vector. Again the parameters of the prior, i.e. $\{v^{(c)}\}_{c=1}^C$, are optimised in the ELBO.
In our case, the input to the inference network is a discrete sentence, which is incompatible with the design of the VampPrior. Thus, we propose to bypass the inference network's embedding layer and estimate a sequence of word embeddings, which makes up a pseudo input. 
That is, $v^{(c)}$ is a sequence $\langle \mathbf v^{(c)}_1, \ldots, \mathbf v^{(c)}_{l_c} \rangle$ where $\mathbf v^{(c)}_i$ has the dimensionality of our embeddings, and $l_c$ is the length of the sequence (fixed at the beginning of training).
Note, however, that for this prior to be multimodal, the inference model must already encode information in $Z$, thus there is some gambling in its design.

\section{\label{sec:experiments}Experiments}

Our goal is to identify which techniques are effective in training VAEs for language modelling. Our evaluation concentrates on intrinsic metrics: negative log-likelihood (NLL), perplexity per token (PPL), rate ($R$), distortion ($D$), the number of active units \citep[AU; ][]{burda2015importance})\footnote{A latent unit (a single dimension of $z$) is denoted \textit{active} when its variance with respect to $x$ is larger than 0.01.} and gap in the accuracy of next word prediction (given gold prefixes) when decoding from a  posterior sample versus decoding from a prior sample (Acc$_\text{gap}$).

For VAE models, NLL (and thus PPL) can only be estimated. We use importance sampling (IS)  
\begin{multline}
\!\!\!\!\!\!P(x|\theta) = \int p(z, x|\theta) \dd z \overset{\text{IS}}{=} \int q(z|x) \frac{p(z, x|\theta)}{q(z|x)} \dd z\\
\!\!\overset{\text{MC}}{\approx} \frac{1}{S} \sum_{s=1}^S \frac{p(z^{(s)}, x|\theta)}{q(z^{(s)}|x)} ~ \text{ where }z^{(s)}\! \sim q(z|x)
\end{multline}
with our trained approximate posterior as importance distribution (we use $S=1000$ samples).

We first report on experiments using the English Penn Treebank  \citep[PTB;][]{marcus1993building}.\footnote{We report on  \citet{dyer2016recurrent}'s pre-processing, rather than \citet{mikolov2010recurrent}'s. Whereas our findings are quantitatively similar, qualitative analysis based on generations are less interesting with Mikolov's far too small vocabulary.}

\paragraph{\textsc{RnnLM}} 
The baseline \textsc{RnnLM} generator is a building block for all of our \textsc{SenVAE}s, thus we validate its performance as a strong standalone generator. 
We highlight that it outperforms an external baseline that employs a comparable number of parameters \citep{dyer2016recurrent} and that this performance boost is mostly due to tying embeddings with the output layer.\footnote{Stronger RNN-based models can be designed \citep{melis2017state}, but those use vastly more parameters.} 
Appendix~\ref{app:architectures} presents the complete architecture and a comparison. %

\paragraph{Bayesian optimisation} The techniques we compare are sensitive to one or more hyperparameters (see Table~\ref{tab:list}), which we tune using Bayesian optimisation (BO) towards minimising estimated NLL of the validation data. For each technique, we ran 25 iterations of BO, each iteration encompassing training a model to full convergence. This was sufficient for the hyperparameters of each technique to converge. See Appendix~\ref{app:BO} for details.

\paragraph{On optimisation strategies}
\begin{table}[t]
\centering
\small
\begin{tabular}{l r}
    \toprule
    Technique & Hyperparameters \\ \midrule 
    KL annealing & increment $\gamma$ ($2\times 10^{-5}$) \\
    Word dropout (WD) & decrement $\gamma$ ($2\times 10^{-5}$)\\
    FB and MDR & target rate $r$ ($5$)\\
    SFB & $r$ ($6.46$), $\gamma$ ($1.05$), $\varepsilon$ ($1$), $\omega$ ($0.01$) \\
    $\beta$-VAE & KL weight $\beta$ ($0.66$) \\
    InfoVAE & $\beta$ ($0.7$), $\lambda$ ($31.62$) \\ 
    LagVAE & $\alpha$ ($-21.7$), target \textsc{MMD} ($0.0017$) \\
    & target -\textsc{ELBO} ($100.8$) \\ \bottomrule
\end{tabular}
\caption{\label{tab:list}Techniques and their hyperparameters.}
\end{table}
\begin{table}[t]
\centering
\small
\begin{tabular}{lHHccHccc}
\toprule
Mode                  & Hyperparameters                   &      $D_p$       &      $D$         &     $R$        & NLL$\downarrow$  &  PPL$\downarrow$ & AU$\uparrow$ & Acc$_\text{gap}$ \\
\midrule
\textsc{RnnLM}        &   -                               &  118.7{\tiny $\pm$0.1} &  -               &  -             &  118.7{\tiny $\pm$0.1} &  107.1{\tiny $\pm$0.5} &  - & -\\
Vanilla               &  -                                &  118.5{\tiny $\pm$0.1} &  118.4 &  0.0 &  118.4{\tiny $\pm$0.1} &  105.7{\tiny $\pm$0.4} & 0 &  0.0 \\
Annealing             & $\gamma$: 2e$^{-5}$               &  127.7{\tiny $\pm$1.0} &  115.3 &  3.3 &  117.9{\tiny $\pm$0.1} &  103.7{\tiny $\pm$0.3} & 17 & 6.0 \\
WD                   & $\gamma$: 2e$^{-5}$               &  117.7{\tiny $\pm$0.2} &  117.6 &  0.0 &  117.6{\tiny $\pm$0.1} &  102.5{\tiny $\pm$0.6} & 0 & 0.0 \\
FB                    & $r$: 5.0                          &  127.0{\tiny $\pm$0.2} &  113.3 &  5.0 &  117.5{\tiny $\pm$0.2} &  101.9{\tiny $\pm$0.8} & 14 &  5.8 \\
SFB                   & $r$: 6.467, $\omega$: 1\%,$\epsilon$: 1,$\gamma$:1.05 & 129.5{\tiny $\pm$0.2} & 112.0 & 6.4 & 117.3{\tiny $\pm$0.1} & 101.0{\tiny $\pm$0.5} & 18 & 7.0 \\
MDR                   & $r$: 5.0                          &  128.0{\tiny $\pm$0.1} &  113.5 &  5.0 &  117.5{\tiny $\pm$0.1} &  102.1{\tiny $\pm$0.5} & 13 & 6.2 \\
$\beta$-VAE           & $\beta$: 0.66                     &  128.5{\tiny $\pm$0.1} &  113.0 &  5.3 &  117.4{\tiny $\pm$0.1} &  101.7{\tiny $\pm$0.5} & 11 & 6.1 \\
InfoVAE               & $\beta$: 0.700, $\lambda$: 31.623 &  126.1{\tiny $\pm$0.1} &  113.5 &  4.3 &  117.2{\tiny $\pm$0.1} &  100.8{\tiny $\pm$0.4} & 10 & 5.2 \\
LagVAE                &                                   &                        &  112.1 &  6.5 &                        &  101.6{\tiny $\pm$0.7} & 24 & 6.9\\
\bottomrule
\end{tabular}
\caption[SenVAE test set results.]{Performance ($\text{avg}${\tiny $\pm \text{std}$} across $5$ independent runs) of \textsc{SenVAE} on the PTB validation set. Standard deviations for $D$ and $R$ are at most $0.2$.} %
\label{tab:opt}
\end{table}
First, we assess the effectiveness of techniques that aim at promoting local optima of \textsc{SenVAE} with better MI tradeoff.
As for the architecture, the approximate posterior $q(z|x, \lambda)$ employs a bidirectional recurrent encoder, and the generator $P(x|z, \theta)$ is essentially our \textsc{RnnLM} initialised with a learned projection of $z$ (complete specification in \ref{app:architectures}).
We train with Adam \citep{kingma2014adam} with default parameters and a learning rate of $10^{-3}$ until convergence five times for each technique.

Results can be found in Table \ref{tab:opt}. First, note how the vanilla VAE (no special treatment) encodes no information in latent space ($R=0$). Then note that all techniques converged to VAEs that attain better PPL than the \textsc{RnnLM} and vanilla VAE, and all but annealed word dropout did so at non-negligible rate. 
Notably, the two most popular techniques, word dropout and $\KL$ annealing, perform sub-par to the other techniques.\footnote{Though here we show annealed word dropout, to focus on techniques that do not weaken the generator, standard word dropout also converged to negligible rates.} 
The techniques that work well at non-negligible rates can be separated into two groups:
one based on a change of objective (i.e., $\beta$-VAE, InfoVAE and LagVAE), another based on targeting a specific rate (i.e., FB, SFB, and MDR).
InfoVAE, LagVAE and SFB all require tuning of multiple hyperparameters. InfoVAE and LagVAE, in particular, showed poor performance without this careful tuning.
In the first group, consider LagVAE, for example. 
Though \citet{zhao2018information} argue that the magnitude of $\alpha$ is not particularly important (in MI-maximisation mode, they fixed it to $-1$), we could not learn a useful \textsc{SenVAE} with LagVAE until we allowed BO to also estimate the magnitude of $\alpha$. Once BO converges to the values in Table~\ref{tab:list}, the method does perform quite well.

\begin{figure}
    \centering
    \includegraphics[scale=0.5]{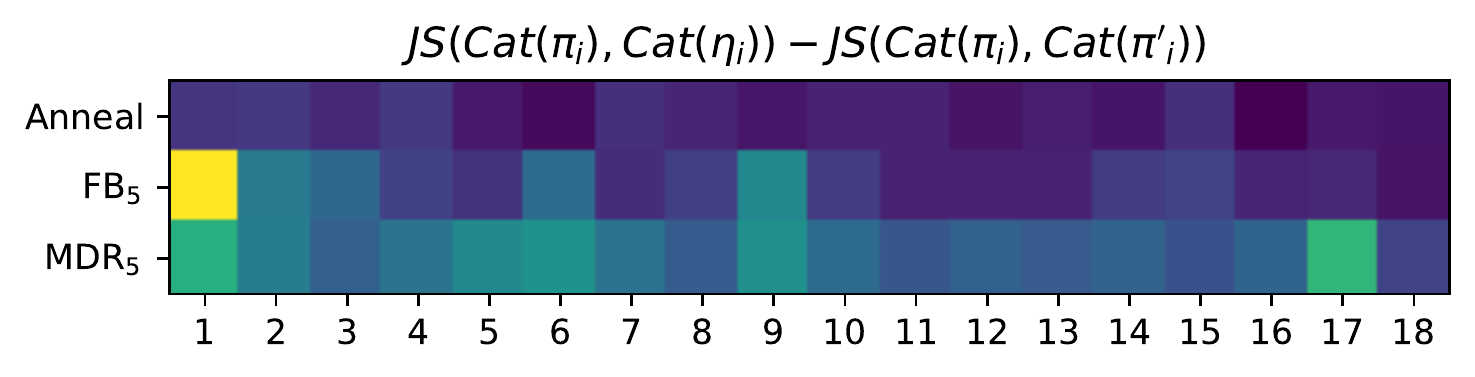}
    \includegraphics[scale=0.5]{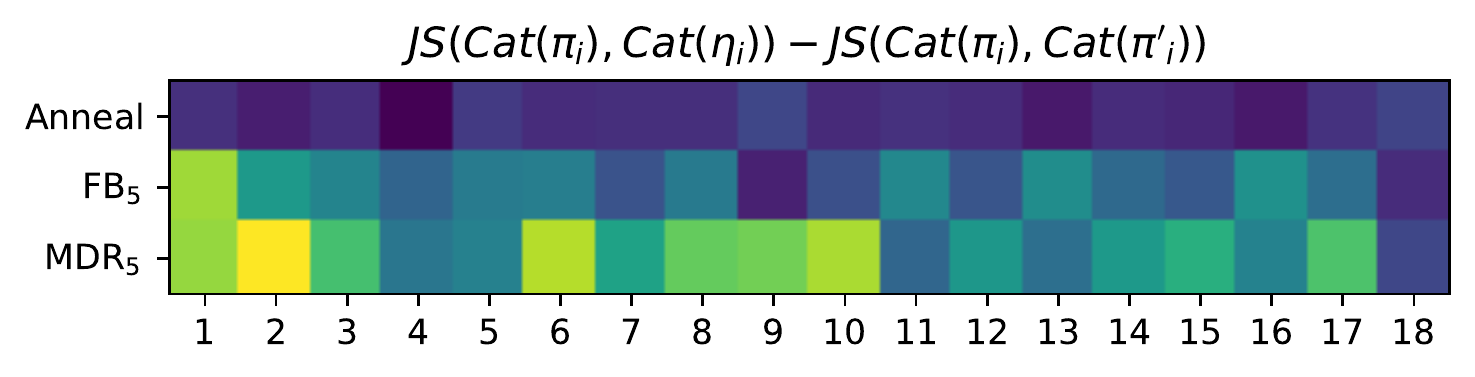}
    \caption{Sensitivity of output distributions to posterior samples measured in terms of symmetrised KL (JS). We obtain $51$ (top) validation and $84$ (bottom) test instances of length $20$ and report on their output distributions per time step. To account for expected variability, we report $\JS(\Cat(\pi_i) || \Cat(\eta_i)) - \JS(\Cat(\pi_i) || \Cat(\pi'_i))$, where $\eta_i$ conditions on a prior sample, and $\pi_i$ and $\pi'_i$ condition on different posterior samples, averaged over $10$ experiments. }
    \label{fig:outputJS}
\end{figure}

Generally, it is hard to believe that hyperparameters transfer across data sets, thus it is fair to expect that this exercise will have to be repeated every time. 
We argue that the rate hyperparameter common to the techniques in the second group is more interpretable and practical in most cases. For example, it is easy to grid-search against a handful of values. %
Hence, we further investigate FB and MDR by varying the target rate further (from $5$ to $50$). SFB is left out, for MDR generalises SFB's handcrafted update rule.
We observe that FB and MDR attain essentially the same PPL across rates, though MDR attains the desired rate earlier on in training, especially for higher targets (where FB fails at reaching the specified rate). Importantly, at the end of training, the validation rate is closer to the target for MDR. Appendix \ref{app:evidence} supports these claims.
Though Acc$_{\text{gap}}$ already suggests it, Figure \ref{fig:outputJS} shows more visibly that MDR leads to output Categorical distributions that are more sensitive to the latent encoding. We measure this sensitivity in terms of symmetrised KL between output distributions obtained from a posterior sample and output distributions obtained from a prior sample for the same time step given an observed prefix.

\paragraph{On expressive priors}
\begin{table}
\centering
\small
\begin{tabular}{lHHHccccc}
\toprule
Model & $q$  &  $p$     &    $D_p$         &    $D$         &    $R$         &  PPL$\downarrow$ &AU$\uparrow$& Acc$_\text{gap}$\\
\hline
\textsc{RnnLM} & -          & -          & 108.5{\tiny $\pm$ 0.2} &  -               &  -             & 84.5{\tiny $\pm$ 0.5} & -            & -              \\
$\mathcal N$/$\mathcal N$ & Gaussian   & Gaussian   & 117.3{\tiny $\pm$ 0.2} & 103.5 & 5.0 & 81.5{\tiny$\pm$ 0.5} & 13 & 5.4 \\
MoG/$\mathcal N$ & Gaussian   & MoG        & 119.7{\tiny $\pm$ 0.2} & 103.3 & 5.0 & 81.4{\tiny$\pm$ 0.5} & 32 & 5.8 \\
Vamp/$\mathcal N$ &Gaussian   & Vamp       & 120.8{\tiny $\pm$ 0.2} & 103.1 & 5.0 & 81.2{\tiny$\pm$ 0.4} & 22 & 5.8 \\
\bottomrule
\end{tabular}
\caption{Performance on the PTB test set for different priors ($\mathcal N$, MoG, Vamp). Standard deviations of $D$, $R$, and Acc$_\text{gap}$ are at most $0.1$.} %
\label{tab:flow}
\end{table}

Second, we compare the impact of expressive priors. 
This time, prior hyperparameters were selected via grid search and can be found in Appendix \ref{app:architectures}. 
All models are trained with a target rate of $5$ using MDR, with settings otherwise the same as the previous experiment.
In Table \ref{tab:flow} it can be seen that more expressive priors do not improve perplexity further,\footnote{Here we remark that best runs (based on validation performance) do show an advantage, which stresses the need to report multiple runs as we do.} though they seem to encode more information in the latent variable---note the increased number of active units and the increased gap in accuracy.
One may wonder whether stronger priors allow us to target higher rates without hurting PPL.
This does not seem to be the case: as we increase rate to $50$, 
 all models perform roughly the same, and beyond $20$ performance degrades quickly.\footnote{We also remark that, without MDR, the MoG model attains validation rate of about $2.5$.}
The models did, however, show a further increase in active units (VampPrior) and accuracy gap (both priors).  %
Again, Appendix \ref{app:evidence} contains plots supporting these claims.

\begin{table*}
\centering
\small
\begin{tabular}{lcccc c cccc}
\toprule
 & \multicolumn{4}{c}{Yahoo} & & \multicolumn{4}{c}{Yelp} \\ \cmidrule{2-5} \cmidrule{7-10}
Model 
 & $R$ 
 & NLL$\downarrow$ 
 & PPL$\downarrow$ 
 & AU$\uparrow$ 
 & 
 & $R$
 & NLL$\downarrow$ 
 & PPL$\downarrow$ 
 & AU$\uparrow$ \\ 
\hline
\textsc{RnnLM}
 & -
 & $328.0$\tiny{$\pm 0.3$} 
 & -
 & -
 & 
 & -
 & $358.1$\tiny{$\pm 0.6$}
 & -
 & - \\
Lagging 
 & $5.7$\tiny{$\pm 0.7$} 
 & $326.7$\tiny{$\pm 0.1$} 
 & -
 & $15.0$\tiny{$\pm 3.5$}
 &
 & $3.8$\tiny{$\pm 0.2$} 
 & $355.9$\tiny{$\pm 0.1$} 
 & -
 & $11.3$\tiny{$\pm 1.0$} \\
$\beta$-VAE ($\beta=0.4$)
 & $6.3$\tiny{$\pm 1.7$} 
 & $328.7$\tiny{$\pm 0.1$} 
 & -
 & $8.0$\tiny{$\pm 5.2$} 
 & 
 & $4.2$\tiny{$\pm 0.4$} 
 & $358.2$\tiny{$\pm 0.3$}
 & -
 & $4.2$\tiny{$\pm 3.8$} \\
Annealing
 & $0.0$\tiny{$\pm 0.0$} 
 & $328.6$\tiny{$\pm 0.0$} 
 & -
 & $0.0$\tiny{$\pm 0.0$} 
 & 
 & $0.0$\tiny{$\pm 0.0$} 
 & $357.9$\tiny{$\pm 0.1$}
 & -
 & $0.0$\tiny{$\pm 0.0$}   
 \\ \midrule
Vanilla 
 & $0.0$\tiny{$\pm 0.0$} 
 & $328.9$\tiny{$\pm 0.1$} 
 & $61.4$\tiny{$\pm 0.1$} 
 & $0.0$\tiny{$\pm 0.0$}
 &
 & $0.0$\tiny{$\pm 0.0$} 
 & $358.3$\tiny{$\pm 0.2$} 
 & $40.8$\tiny{$\pm 0.1$} 
 & $0.0$\tiny{$\pm 0.0$}  \\
$\mathcal N$/$\mathcal N$ 
 & $5.0$\tiny{$\pm 0.0$} 
 & $328.1$\tiny{$\pm 0.1$} 
 & $60.8$\tiny{$\pm 0.1$}
 & $4.0$\tiny{$\pm 0.7$}
 & 
 & $5.0$\tiny{$\pm 0.0$} 
 & $357.5$\tiny{$\pm 0.2$}
 & $40.4$\tiny{$\pm 0.1$}
 & $4.2$\tiny{$\pm 0.4$}\\ 
MoG/$\mathcal N$
 & $5.0$\tiny{$\pm 0.1$} 
 & $327.5$\tiny{$\pm 0.2$}
 & $60.5$\tiny{$\pm 0.1$}
 & $5.0$\tiny{$\pm 0.7$}
 & 
 & $5.0$\tiny{$\pm 0.0$} 
 & $359.5$\tiny{$\pm 0.5$}
 & $41.2$\tiny{$\pm 0.3$}
 & $2.2$\tiny{$\pm 0.4$}\\ 
\bottomrule
\end{tabular}
\caption{Performance on the Yahoo/Yelp corpora. Top rows taken from \citep{he2018lagging}}
\label{tab:largedata}
\end{table*}

\paragraph{Generated samples}
\begin{figure*}[t]
\small
\centering
\begin{tabular}{@{}p{0.45\textwidth}p{0.45\textwidth}r@{}}
\toprule
Sample & Closest training instance & TER \\ \midrule
For example, the Dow Jones Industrial Average fell almost 80 points to close at 2643.65.                           
& \textit{By futures-related program buying, the Dow Jones Industrial Average gained 4.92 points to close at 2643.65.} 
& $0.38$ \\ \midrule
The department store concern said it expects to report profit from continuing operations in 1990.  
& \textit{Rolls-Royce Motor Cars Inc. said it expects its U.S. sales to remain steady at about 1,200 cars in 1990.}
& $0.59$ \\ \midrule 
The new U.S. auto makers say the accord would require banks to focus on their core businesses of their own account.                                                 
& \textit{International Minerals said the sale will allow Mallinckrodt to focus its resources on its core businesses of medical products, specialty chemicals and flavors.}
& $0.78$ \\ \bottomrule
\end{tabular}
\caption{Samples from \textsc{SenVAE} (MoG prior) trained via MDR: we sample from the prior and decode greedily. We also show the closest training instance in terms of a string edit distance (TER).}
\label{fig:good-prior-samples}
\end{figure*}

\begin{figure}[]
\small
\fontdimen2\font=1pt
\begin{tabular}{@{}l@{}}
\toprule
\textbf{The inquiry soon focused on the judge.} \\
The judge declined to comment on the floor. \\
The judge was dismissed as part of the settlement. \\
The judge was sentenced to death in prison. \\
The announcement was filed against the SEC. \\
The offer was misstated in late September. \\
 The offer was filed against bankruptcy court in New York. \\
\textbf{The letter was dated Oct. 6.}\\
\bottomrule
\end{tabular}
\caption{Latent space homotopy from a properly trained \textsc{SenVAE}. Note the smooth transition of topic and grammatically of the samples.}
\label{fig:good-hom}
\end{figure}

Figure~\ref{fig:good-prior-samples} shows samples from a well-trained \textsc{SenVAE}, where we decode greedily from a prior sample---this way, all variability is due to the generator's reliance on the latent sample. Recall that a vanilla VAE ignores $z$ and thus greedy generation from a prior sample is essentially deterministic in that case (see Figure~\ref{fig:vanilla-samples}). 
Next to the samples we show the closest training instance, which we measure in terms of an edit distance \citep[TER; ][]{snover2006study}.\footnote{This distance metric varies from $0$ to $1$, where $1$ indicates the sentence is completely novel and $0$ indicates the sentence is essentially copied from the training data.}
This ``nearest neighbour'' helps us assess whether the generator is producing novel text or simply reproducing something it memorised from training. 
In Figure~\ref{fig:good-hom} we show a homotopy: here we decode greedily from points lying between a posterior sample conditioned on the first sentence and a posterior sample conditioned on the last sentence.
In contrast to the vanilla VAE (Figure~\ref{fig:vanilla-hom}), neighbourhood in latent space is  now used to capture some regularities in data space. 
These samples add support to the quantitative evidence that our DGMs have been trained not to neglect the latent space. In Appendix~\ref{app:evidence} we provide more samples.  %

\paragraph{Other datasets} To address the generalisability of our claims to other, larger, datasets, we report results on the Yahoo and Yelp corpora \cite{yang2017improved} in Table \ref{tab:largedata}. We compare to the work of \citet{he2018lagging}, who proposed to mitigate posterior collapse with aggressive training of the inference network, optimising the inference network multiple steps for each step of the generative network.\footnote{To enable direct comparison we replicated the experimental setup from \cite{he2018lagging} and built our methods into their codebase.}. 
We report on models trained with the standard prior as well as an MoG prior both optimised with MDR, and a model trained without optimisation techniques.\footnote{We focus on MoG since the PTB experiments showed the VampPrior to underperform in terms of AU.} It can be seen that MDR compares favourably to other optimisation techniques reported in \cite{he2018lagging}. Whilst aggressive training of the inference network performs slightly better in terms of NLL and leads to more active units, it slows down training by a factor of 4. The MoG prior improves results on Yahoo but not on Yelp. This may indicate that a multimodal prior does offer useful extra capacity to the latent space,\footnote{We tracked the average KL divergence between any two components of the prior and observed that the prior remained multimodal.} at the cost of more instability in optimisation.
This confirms that targeting a pre-specified rate leads to VAEs that are not collapsed without hurting NLL.

\paragraph{Recommendations} We recommend targeting a specific rate via MDR instead of annealing (or word dropout). Besides being simple to implement, it is fast and straightforward to use: pick a rate by plotting validation performance against a handful of values. 
Stronger priors, on the other hand, while showing indicators of higher mutual information (e.g. AU and Acc$_{\text{gap}}$), seem less effective than MDR. 
Use IS estimates of NLL, rather than single-sample ELBO estimates, for model selection, for the latter can be too loose of a bound and too heavily influenced by noisy estimates of KL.\footnote{This point seems obvious to many, but enough published papers report exponentiated loss or distortion per token, which, besides  unreliable, make comparisons across papers difficult.} 
Use many samples for a tight bound.\footnote{We use $1000$ samples. Compared to a single sample estimate, we have observed differences up to $5$ perplexity points in non-collapsed models. From $100$ to $1000$ samples, differences are in the order of $0.1$ suggesting our IS estimate is close to convergence.}
Inspect sentences greedily decoded from a prior (or posterior) sample as this shows whether the generator is at all sensitive to the latent code. Retrieve nearest neighbours to spot copying behaviour. %

\section{\label{sec:related}Related Work}

In NLP, posterior collapse was probably first noticed by  \citet{bowman2016generating}, who addressed it via word dropout and KL scaling.
Further investigation revealed that in the presence of strong generators, the ELBO itself becomes the culprit \citep{chen2016variational, alemi2018fixing}, as it lacks a preference regarding MI. %
Posterior collapse has also been ascribed to approximate inference \citep{kim2018semi,dieng2019}.
Beyond the techniques compared and developed in this work, other solutions have been proposed, including modifications to the generator  \citep{semeniuta2017hybrid, yang2017improved, park2018dialogue, Dieng2018AvoidingLV}, side losses based on weak generators \citep{zhao2017learning}, factorised likelihoods \citep{ziegler2019latent,ma-etal-2019-flowseq}, cyclical annealing \citep{liu2019cyclical} and changes to the ELBO \citep{tolstikhin2017wasserstein, goyal2017z}.

Exploiting a mismatch in correlation between the prior and the approximate posterior, and thus forcing a lower-bound on the rate, is the principle behind $\delta$-VAEs \citep{razavi2019preventing} and hyperspherical VAEs \citep{xu2018spherical}. The generative model of $\delta$-VAEs has one latent variable per step of the sequence, i.e. $z = \langle z_1, \ldots, z_{|x|} \rangle$, making it quite different from that of the \textsc{SenVAE}s considered here. Their mean-field inference model is a product of independent Gaussians, one per step, but they construct a correlated Gaussian prior by making the prior distribution over the next step depend linearly on the previous step, i.e. $Z_i|z_{i-1} \sim \mathcal N(\alpha z_{i-1}, \sigma)$ with hyperparameters $\alpha$ and $\sigma$. 
Hyperspherical VAEs work on the unit hypersphere with a uniform prior and a non-uniform VonMises-Fisher posterior approximation \citep{davidson2018hyperspherical}.
Note that, though in this paper we focused on Gaussian (and mixture of Gaussians, e.g. MoG and VampPrior) priors, MDR is applicable for whatever choice of prescribed prior. Whether its benefits stack with the effects due to different priors remains an empirical question.

GECO \citep{rezende2018taming} casts VAE optimisation as a dual problem, and in that it is closely related to our MDR and the LagVAE.
GECO targets minimisation of $\mathbb E_X[\KL(q(z|x,\lambda)||p(z))]$ under constraints on distortion, whereas LagVAE targets either maximisation or minimisation of (bounds on) $I(X;Z|\lambda)$ under constraints on the InfoVAE objective. 
Contrary to MDR, GECO focuses on latent space regularisation and offers no explicit mechanism to mitigate posterior collapse.

Recently \citet{li-etal-2019-surprisingly} proposed to combine FB, KL scaling, and pre-training of the inference network's encoder on an auto-encoding objective. Their techniques are complementary to ours in so far as their main finding---the mutual benefits of annealing, pre-training, and lower-bounding KL---is perfectly compatible with ours (MDR and multimodal priors). %

\section{Discussion}

\textsc{SenVAE} is a deep generative model whose generative story is rather shallow, yet, due to its strong generator component, it is hard to make effective use of the extra knob it offers.
In this paper, we have introduced and compared techniques for effective estimation of such a model. We show that many techniques in the literature perform reasonably similarly (i.e. FB, SFB, $\beta$-VAE, InfoVAE), though they may  require a considerable hyperparameter search (e.g. SFB and InfoVAE). 
Amongst these, our proposed optimisation subject to a minimum rate constraint is simple enough to tune (as FB it only takes a pre-specified rate and unlike FB it does not suffer from gradient discontinuities), superior to annealing and word dropout, and require less resources than strategies based on multiple annealing schedules and/or aggressive optimisation of the inference model. 
Other ways to lower-bound rate, such as by imposing a multimodal prior, though promising, still require a minimum desired rate. 

The typical \textsc{RnnLM} is built upon an exact factorisation of the joint distribution, thus a well-trained architecture is hard to improve upon in terms of log-likelihood of gold-standard data. 
Our interest in latent variable models stems from the desire to obtain generative stories that are less opaque than that of an \textsc{RnnLM}, for example, in that they may expose knobs that we can use to control generation and a hierarchy of steps that may award a degree of interpretability to the model. The \textsc{SenVAE} \emph{is not} that model, but \emph{it is} a crucial building block in the pursue for hierarchical probabilistic models of language. 
We hope this work, i.e. the organised review it contributes and the techniques it introduces, will pave the way to deeper---in \emph{statistical hierarchy}---generative models of language.

\section*{Acknowledgments}

This project has received funding from the European Union's Horizon 2020 research and innovation programme under grant agreement No 825299  (GoURMET).

\bibliography{acl2020}
\bibliographystyle{acl_natbib}

\clearpage

\appendix

\section{Architectures and Hyperparameters}
In order to ensure that all our experiments are fully reproducible, this section provides an extensive overview of the model architectures, as well as model and optimisation hyperparameters.

Some hyperparameters are common to all experiments, e.g. optimiser and dropout, they can be found in Table \ref{tab:expparam}. All models were optimised with Adam using default settings \citep{kingma2014adam}. To regularise the models, we use dropout with a shared mask across time-steps \citep{zaremba2014recurrent} and weight decay proportional to the dropout rate \citep{gal2015dropout} on the input and output layers of the generative networks (i.e. \textsc{RnnLM} and the recurrent decoder in \textsc{SenVAE}). 
No dropout is applied to layers of the inference network as this does not lead to consistent empirical benefits and lacks a good theoretical basis. 
Gradient norms are clipped to prevent exploding gradients, and long sentences are truncated to three standard deviations above the average sentence length in the training data.  %

\begin{table}[h]
\centering
\small
\begin{tabular}{@{}ll@{}}
\toprule
Parameter                     & Value        \\ \midrule
Optimizer                     & Adam        \\
Optimizer Parameters          & $\beta_1=0.9, \beta_2=0.999$ \\ 
Learning Rate                 & 0.001        \\
Batch Size                    & 64           \\
Decoder Dropout Rate ($\rho$) & 0.4          \\
Weight Decay                  & $\frac{1-\rho}{|\mathcal D|}$ \\
Maximum Sentence Length       & 59           \\
Maximum Gradient Norm         & 1.5          \\ \bottomrule
\end{tabular}
\caption{Experimental settings.}
\label{tab:expparam}
\end{table}

\subsection{\label{app:architectures}Architectures}

\begin{table}[t]
\centering
\small
\begin{tabular}{@{}lll@{}}
\toprule
Model & Parameter                   & Value        \\ \midrule
A & embedding units ($\dims_e$)                 & 256          \\
A    & vocabulary size ($\dims_v$)                 & 25643        \\
R and S & decoder layers ($L^{\theta}$)               & 2            \\
R and S & decoder hidden units ($\dims_h^{\theta}$)   & 256          \\
S & encoder hidden units ($\dims_h^{\lambda}$)     & 256          \\
S & encoder layers ($L^{\lambda}$)                 & 1            \\
S & latent units    ($\dims_z$)                 & 32           \\
MoG & mixture components ($C$) & 100 \\
VampPrior & pseudo inputs ($C$) & 100 \\
\bottomrule
\end{tabular}
\caption{Architecture parameters: all (A), \textsc{RnnLM} (R), \textsc{SenVAE} (S).} %
\label{tab:structparam}
\end{table}

This section describes the components that parameterise our models.\footnote{All models were implemented with the \textsc{PyTorch} library \citep{paszke2017automatic}, using default modules for the recurrent networks, embedders and optimisers.} 
We use mnemonic blocks $\layer(\text{inputs}; \text{parameters})$ to describe architectures.
Table \ref{tab:structparam} lists  hyperparameters for the models discussed in what follows.

\paragraph{\textsc{RnnLM}}
At each step, an \textsc{RnnLM} parameterises a categorical distribution over the vocabulary, i.e. $X_i| x_{<i} \sim \Cat(f(x_{<i}; \theta))$, where $f(x_{<i}; \theta) = \softmax(\mat o_{i})$ and
\begin{subequations}
\begin{align}
  \mathbf e_i &= \emb(x_i; \theta_{\text{emb}}) \\
  \mathbf h_{i} &= \GRU(\mathbf h_{i-1}, \mathbf e_{i-1}; \theta_{\text{gru}})\\
  \mathbf o_i &= \affine(\mathbf h_i; \theta_{\text{out}})  ~ .
\end{align}
\end{subequations}
We employ an embedding layer ($\emb$), one (or more) $\GRU$ cell(s) ($\mathbf h_0 \in \theta$ is a parameter of the model), and an $\affine$ layer to map from the dimensionality of the GRU to the vocabulary size. 
Table~\ref{tab:det-base} compares our \textsc{RnnLM} to an external baseline with a comparable number of parameters.
\begin{table}[h]
\centering
    \begin{tabular}{@{}llll@{}}
    
    \toprule
    Model   & PPL$\downarrow$ & PPL$^{\text{Dyer}}\downarrow$ \\ \midrule
    \citet{dyer2016recurrent} & $93.5$        & $113.4$                \\
    \textsc{RnnLM}        & 84.5{\tiny $\pm$ 0.52} & $102.1$                \\ \bottomrule
    \end{tabular}
\caption[\textsc{RnnLM} test set results.]{Baseline LMs on the PTB test set: $\text{avg} \pm \text{std}$ over $5$ independent runs. Unlike us, \citet{dyer2016recurrent} removed the end of sentence token for evaluation, thus the last column reports perplexity computed that way.} %
\label{tab:det-base}
\end{table}

\paragraph{Gaussian \textsc{SenVAE}}
A Gaussian \textsc{SenVAE} also parameterises a categorical distribution over the vocabulary for each given prefix, but, in addition, it conditions on a latent embedding $Z \sim \mathcal N(0, I)$, i.e. $X_i|z, x_{<i} \sim \Cat(f(z, x_{<i}; \theta))$ where $f(z, x_{<i}; \theta) = \softmax(\mat o_{i})$ and
\begin{subequations}
\begin{align}
  \mathbf e_i &= \emb(x_i; \theta_{\text{emb}}) \\
  \mathbf h_0 &= \tanh(\affine(z; \theta_{\text{init}})) \\
  \mathbf h_{i} &= \GRU(\mathbf h_{i-1}, \mathbf e_{i-1}; \theta_{\text{gru}})\\
  \mathbf o_i &= \affine(\mathbf h_i; \theta_{\text{out}}) ~ .
\end{align}
\end{subequations}
Compared to \textsc{RnnLM}, we modify $f$ only slightly by initialising GRU cell(s) with $\mathbf h_0$ computed as a learnt transformation of $z$.
Because the marginal of the Gaussian \textsc{SenVAE} is intractable, we train it via variational inference using an inference model $q(z|x, \lambda) = \mathcal N(z|\mathbf u, \diag(\mathbf s \odot \mathbf s))$ where
\begin{subequations}
\begin{align}
    \mathbf e_i &= \emb(x_i; \theta_{\text{emb}}) \\
    \mathbf h_1^n &= \BiGRU(\mat e_1^n, \mat h_0; \lambda_{\text{enc}})\\
    \mathbf u &= \affine(\mathbf h_n; \lambda_{\text{loc}}) \\
    \mathbf s &= \softplus(\affine(\mathbf h_n; \lambda_{\text{scale}})) ~ .
\end{align}
\end{subequations}
Note that we reuse the embedding layer from the generative model. Finally, a sample is obtained via $z = \mathbf u + \mathbf s \odot \epsilon $ where $\epsilon \sim \mathcal N(0, I_{\dims_z})$.

\paragraph{MoG prior} We parameterise $C$ diagonal Gaussians, which are mixed uniformly. To do so we need $C$ location vectors, each in $\mathbb R^{\dims_z}$, and $C$ scale vectors, each in $\mathbb R^{\dims_z}_{>0}$. To ensure strict positivity for scales we make $\boldsymbol\sigma^{(c)} = \softplus(\hat{\boldsymbol \sigma}^{(c)})$. The set of generative parameters $\theta$ is therefore extended with $\{\boldsymbol \mu^{(c)}\}_{c=1}^C$ and $\{\hat{\boldsymbol \sigma}^{(c)}\}_{c=1}^C$, each in $\mathbb R^{\dims_z}$. 

\paragraph{VampPrior} For this we estimate $C$ sequences $\{v^{(c)}\}_{c=1}^C$ of input vectors, each sequence $v^{(c)} = \langle \mathbf v_1^{(c)}, \ldots, \mathbf v_{l_k}^{(c)} \rangle$ corresponds to a \emph{pseudo-input}. 
This means we extend the set of generative parameters $\theta$ with $\{ \mathbf v_i^{(c)}\}_{i=1}^{l_c}$, each in $\mathbb R^{\dims_e}$, for $c=1, \ldots, C$. For each $c$, we sample $l_c$ at the beginning of training and keep it fixed. Specifically, we drew $C$ samples from a normal, $l_c \sim \mathcal N(\cdot|\mu_l, \sigma_l)$, which we rounded to the nearest integer. $\mu_l$ and $\sigma_l$ are the dataset sentence length mean and variance respectively.

\subsection{\label{app:BO}Bayesian optimisation}

\begin{table}[t]
\centering
\begin{tabular}{@{}ll@{}}
\toprule
Parameter                                   & Value        \\ \midrule
Objective Function                          & Validation NLL        \\
Kernel                                      & Matern$_{52}$          \\
Acquisition Function                        & Expected Improvement   \\
Parameter Inference                         & MCMC                    \\
MCMC Samples                                & 10                      \\
Leapfrog Steps                              & 20                      \\
Burn-in Samples                              & 100                     \\
\bottomrule
\end{tabular}
\caption{Bayesian optimisation settings.}
\label{tab:bayesparam}
\end{table}

Bayesian optimisation (BO) is an efficient method to approximately search for global optima of a (typically expensive to compute) objective function $y = f(\mat x)$, where $\mat x \in \R^{M}$ is a vector containing the values of $M$ hyperparameters that may influence the outcome of the function \citep{snoek2012practical}. Hence, it forms an alternative to grid search or random search \citep{bergstra2012random} for tuning the hyperparameters of a machine learning algorithm. 
BO works by assuming that our observations $y_n|\mathbf x_n$ (for $n=1, \ldots, N$) are drawn from a Gaussian process \citep[GP;][]{RasmussenEtAlGP}. 
Then based on the GP posterior, we can design and infer an acquisition function. This acquisition function can be used to determine where to ``look next'' in  parameter-space, i.e. it can be used to draw $\mat x_{N+1}$ for which we then evaluate the objective function $f(x_{N+1})$. 
This procedure iterates until a set of optimal parameters is found with some level of confidence.

In practice, the efficiency of BO hinges on multiple choices, such as the specific form of the acquisition function, the covariance matrix (or kernel) of the GP and how the parameters of the acquisition function are estimated. 
Our objective function is the (importance-sampled) validation NLL, which can only be computed after a model convergences (via gradient-based optimisation of the ELBO).
We follow the advice of \citet{snoek2012practical} and use MCMC for estimating the parameters of the acquisition function.  
This reduced the amount of objective function evaluations, speeding up the overall search. 
Other settings were also based on results by \citet{snoek2012practical}, and we refer the interested reader to that paper for more information about BO in general. 
A summary of all relevant settings of BO can be found in Table \ref{tab:bayesparam}. We used the \textsc{GPyOpt} library \citep{gpyopt2016} to implement this procedure.

\section{\label{app:evidence}Additional Empirical Evidence}

\begin{figure*}[t]
    \centering
   \begin{subfigure}[t]{0.3\textwidth}
        \includegraphics[scale=0.45]{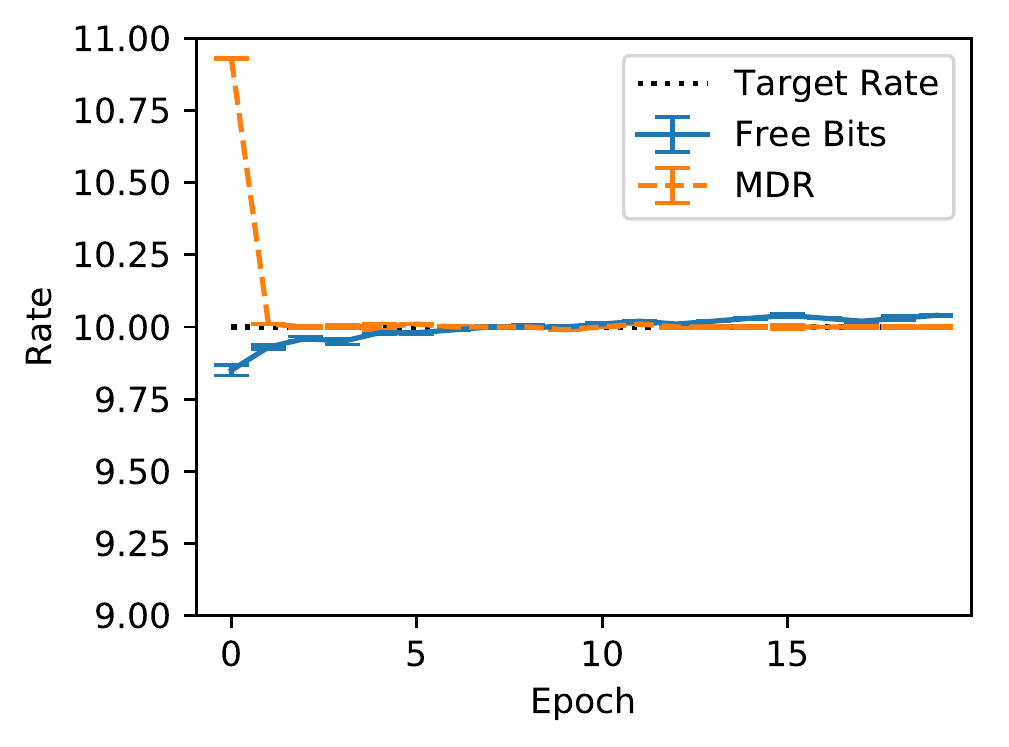}
        \caption{Rate over time for $r=10$.}
    \end{subfigure}
    ~
    \begin{subfigure}[t]{0.3\textwidth}
        \includegraphics[scale=0.45]{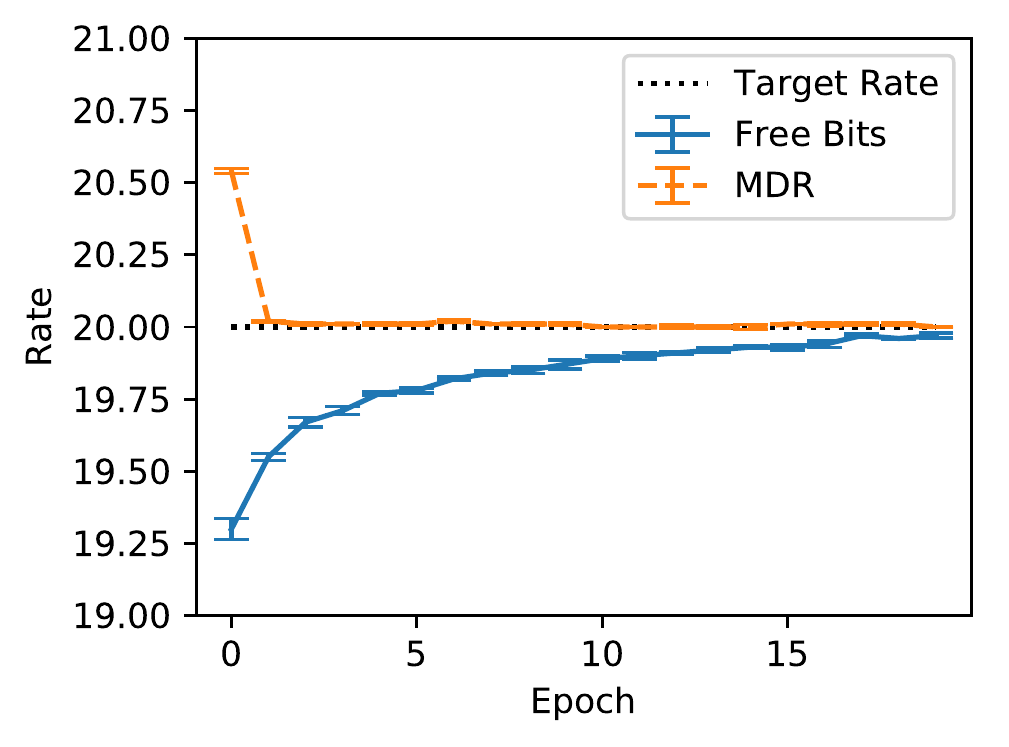}
        \caption{Rate over time for $r=20$.}
    \end{subfigure}
    ~
    \begin{subfigure}[t]{0.3\textwidth}
        \includegraphics[scale=0.45]{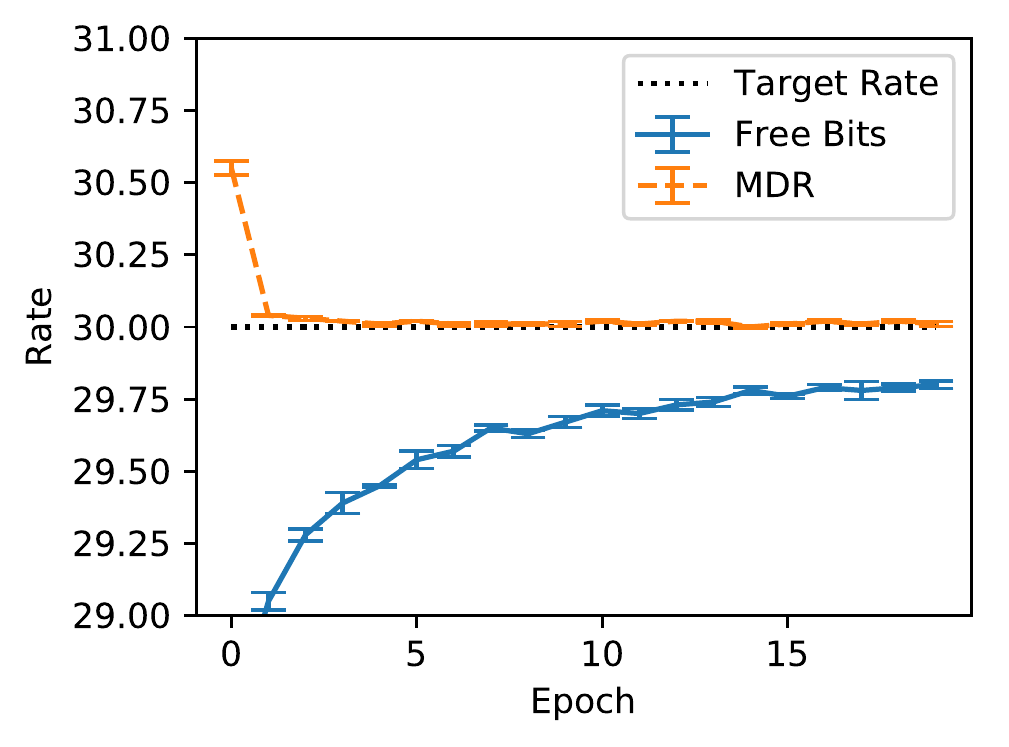}
        \caption{Rate over time for $r=30$.}
    \end{subfigure}
    \caption{ \label{fig:rateprog} Rate progression on the training set in the first $20$ epochs of training for \textsc{SenVAE} trained with free bits (FB) or minimum desired rate (MDR). One can observe that at higher rates, FB struggles to achieve the target rate, whereas MDR achieves the target rate after a few epochs.}
\end{figure*}

\begin{figure*}[t]
    \centering
    \begin{subfigure}[t]{0.45\textwidth}
        \includegraphics[scale=0.45]{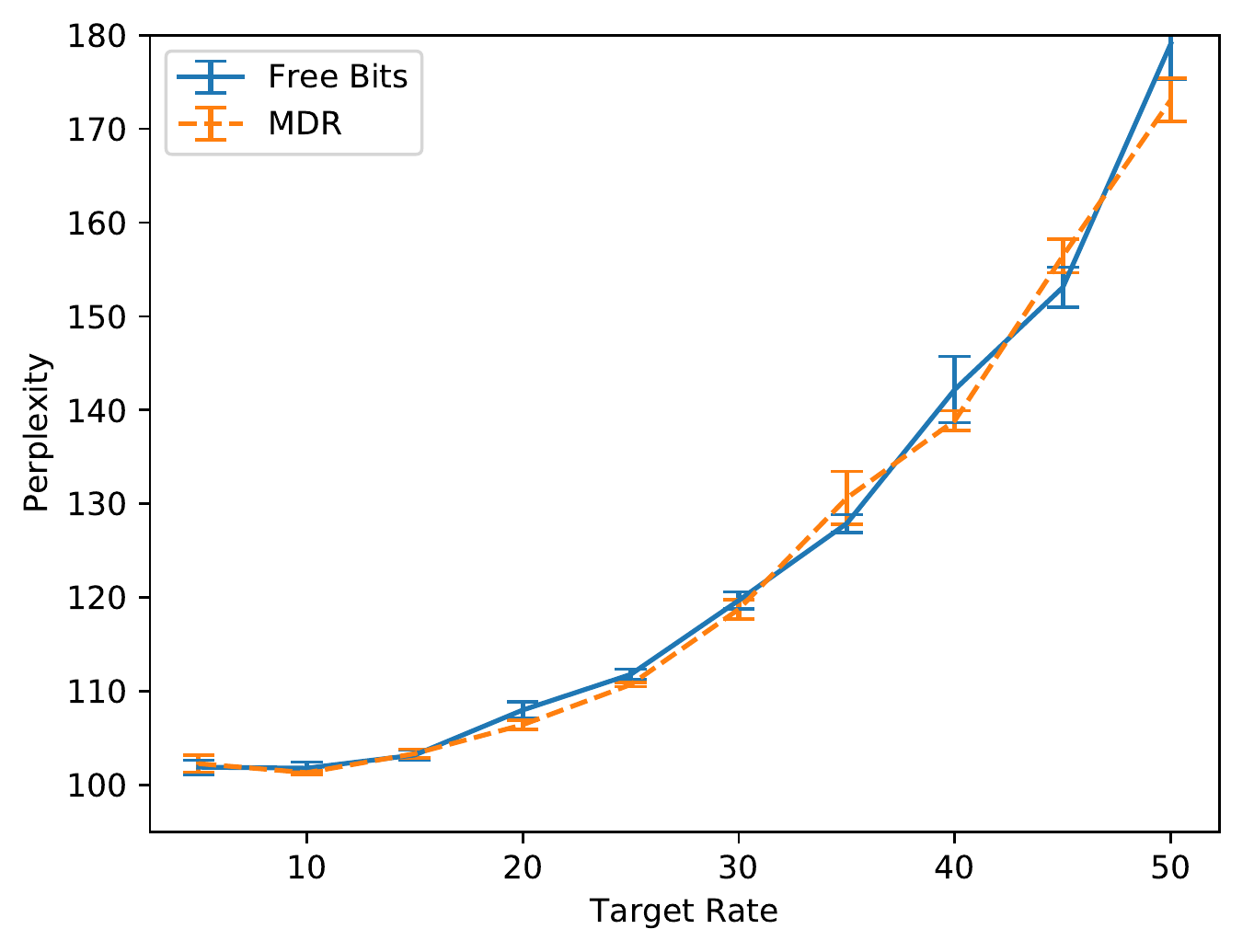}
        \caption{PPL ($\downarrow$) for various target rates.}
        \label{fig:ppl-per-r}
    \end{subfigure}
    ~
    \begin{subfigure}[t]{0.45\textwidth}
        \includegraphics[scale=0.45]{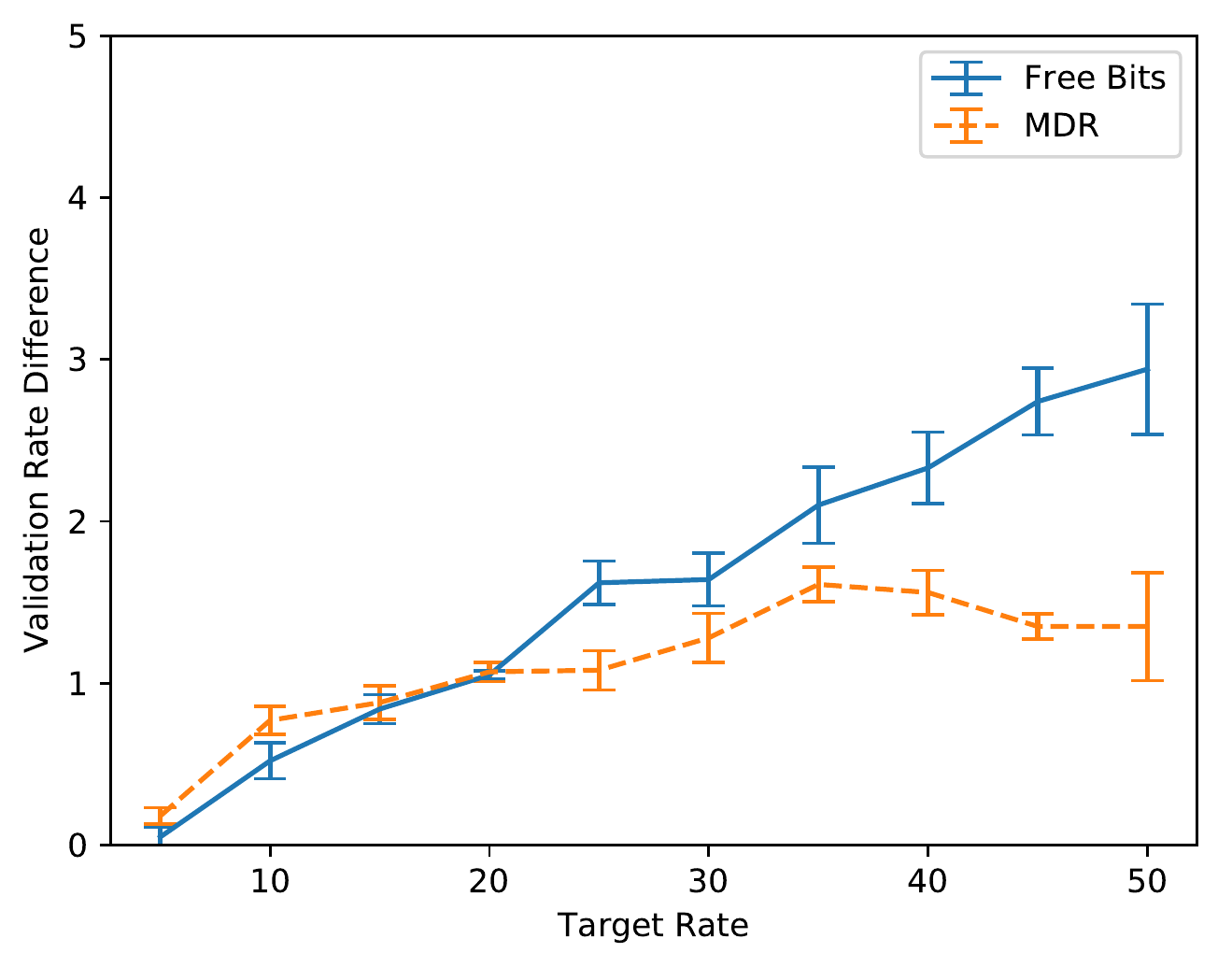}
        \caption{\label{fig:ratediff}Target rate minus validation rate at the end of training for various targets (lower means better).}
    \end{subfigure}
    \caption{\label{fig:mdrisawesome} Validation results for \textsc{SenVAE} trained with free bits (FB) or minimum desired rate (MDR).}
\end{figure*}

In Figure \ref{fig:rateprog} we inspect how MDR and FB approach different target rates (namely, $10$, $20$, and $30$). 
Note how MDR does so more quickly, especially at higher rates.
Figure \ref{fig:ppl-per-r}  shows that in terms of validation perplexity, MDR and FB perform very similarly across target rates. 
However, Figure \ref{fig:ratediff} shows that at the end of training the difference between the target rate and the validation rate is smaller for MDR.

\begin{figure*}[t]
    \centering
    \begin{subfigure}[t]{0.45\textwidth}
        \includegraphics[scale=0.45]{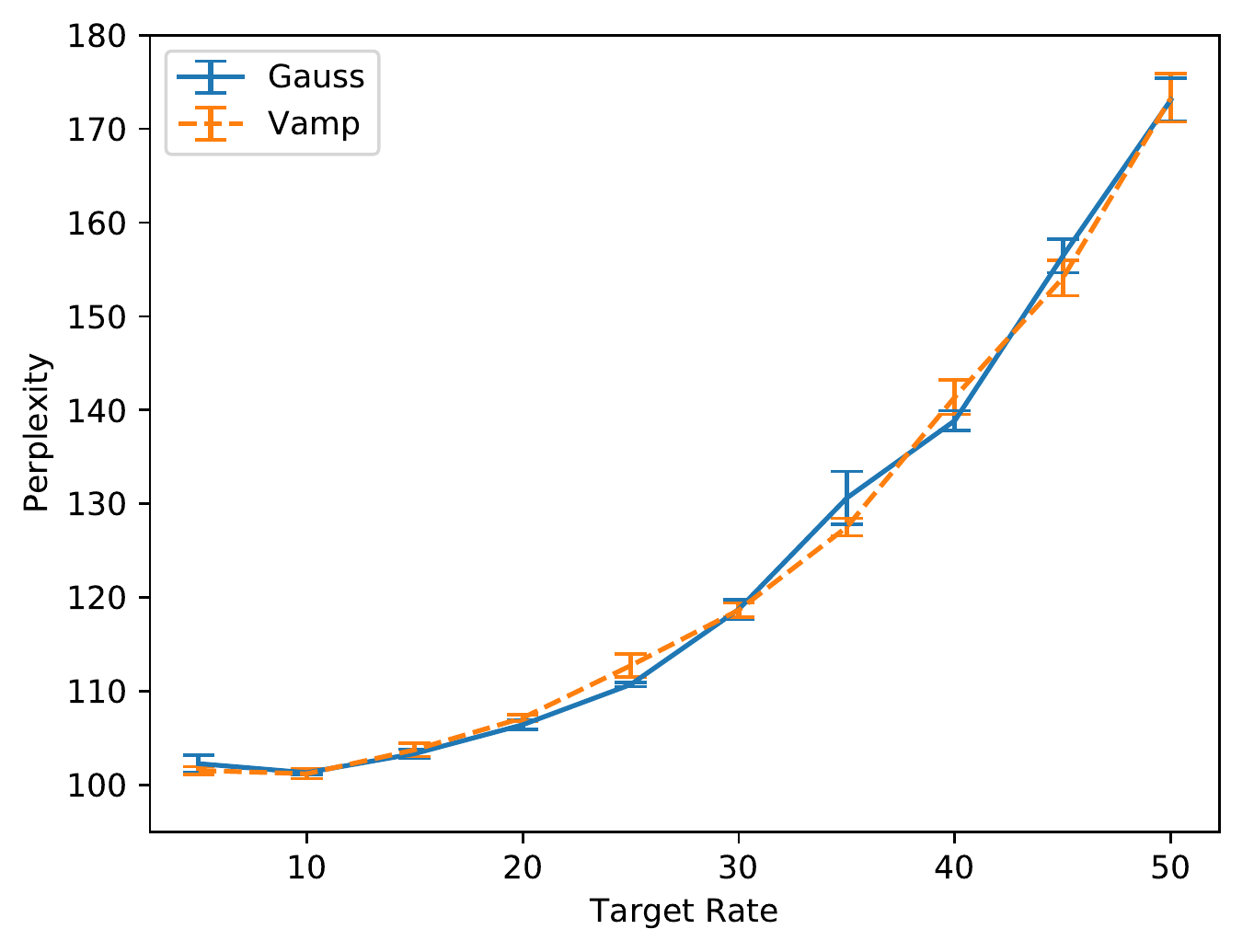}
        \caption{\label{fig:ppl_flow}Perplexity on validation set: models perform similarly well and perplexity degrades considerably for $r > 20$.}
    \end{subfigure}
    ~
    \begin{subfigure}[t]{0.45\textwidth}
        \includegraphics[scale=0.45]{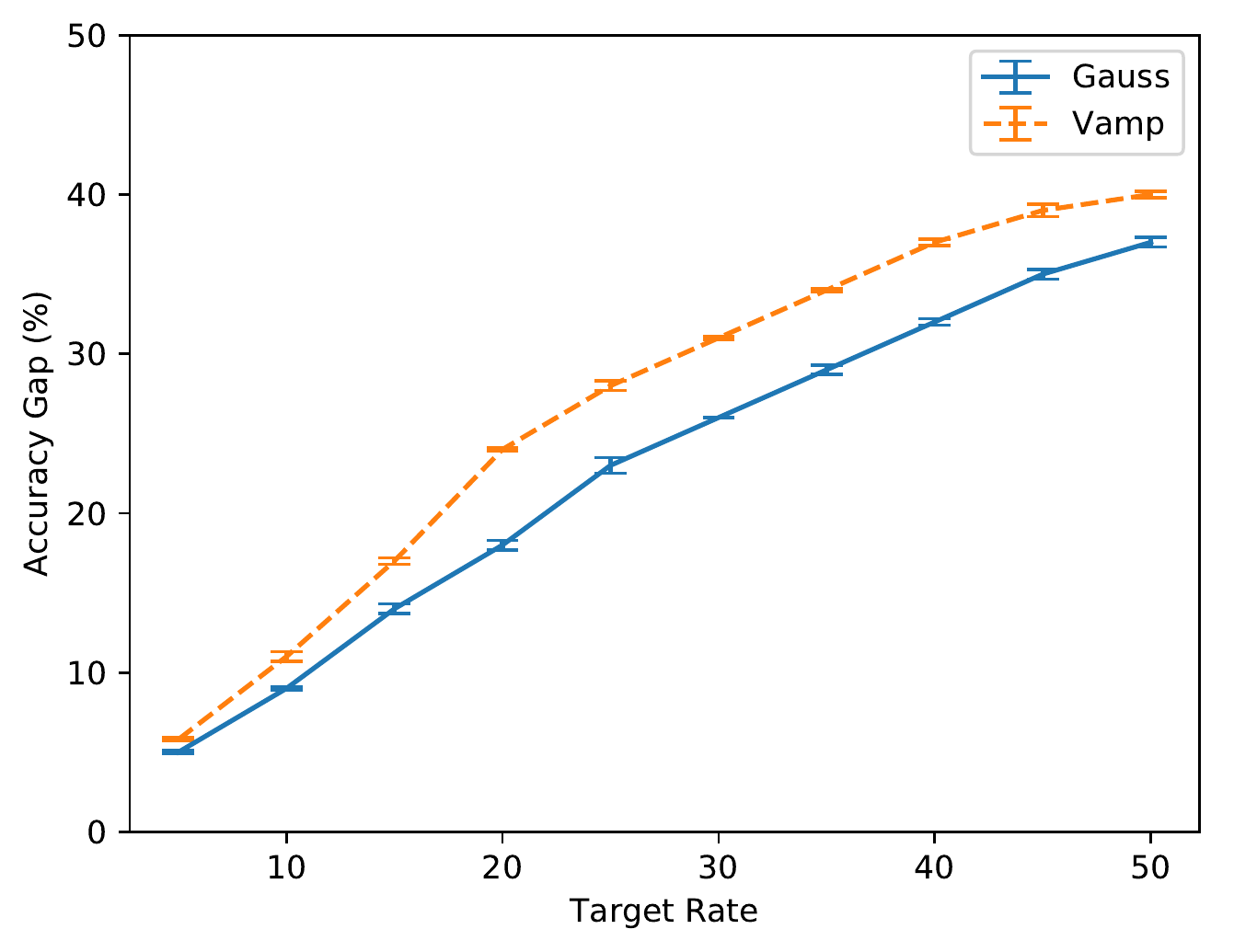}
        \caption{Accuracy gap: VAEs with stronger latent components rely more on posterior samples for reconstruction.}
        \label{fig:dgap_flow}
    \end{subfigure}
    \caption{\label{fig:flow} Comparison of \textsc{SenVAE}s trained with standard prior and Gaussian posterior (\textcolor{blue}{Gauss}) and Vamp prior and Gaussian posterior (\textcolor{orange}{Vamp}) to attain pre-specified rates.}
    
\end{figure*}

Figure~\ref{fig:flow} compares variants of \textsc{SenVAE} trained with MDR for various rates: a Gaussian-posterior and Gaussian-prior (blue-solid) to a Gaussian-posterior and Vamp-prior (orange-dashed).
They perform essentially the same in terms of perplexity (Figure~\ref{fig:ppl_flow}), but the variant with the stronger prior relies more on posterior samples for reconstruction (Figure~\ref{fig:dgap_flow}).

Finally, we list additional samples: Figure~\ref{fig:app-samples} lists samples from \textsc{RnnLM}, vanilla \textsc{SenVAE} and effectively trained variants (via MDR with target rate $r=10$); Figure~\ref{fig:app-hom} lists homotopies from \textsc{SenVAE} models.

\begin{figure*}[t]
\small
\centering
\begin{tabular}{@{}l p{0.38\textwidth}p{0.38\textwidth}r@{}}
\toprule
Model & Sample & Closest training instance & TER \\ \midrule
\multirow{3}{*}{\textsc{RnnLM}} & The Dow Jones Industrial Average jumped 26.23 points to 2662.91 on 2643.65.
& \textit{The Dow Jones Industrial Average fell 26.23 points to 2662.91.} 
& $0.23$ \\ \cmidrule{2-4}
& The companies said they are investigating their own minds with several carriers, including the National Institutes  of Health and Human Services Department of Health,,
& \textit{The Health and Human Services Department currently forbids the National Institutes of Health from funding abortion research as part of its \$8 million contraceptive program.}
& $0.69$ \\ \cmidrule{2-4}
& And you'll have no longer sure whether you would do anything not -- if you want to get you don't know what you're, 
& \textit{Reaching for that extra bit of yield can be a big mistake -- especially if you don't understand what you're investing.} 
& $0.81$ \\  \midrule
\multirow{3}{*}{\textsc{SenVAE}} & The company said it expects to report net income of \$UNK-NUM million, or \$1.04 a share, from \$UNK-NUM  million, or,
& \textit{Nine-month net climbed 19\% to \$UNK-NUM million, or \$2.21 a primary share, from \$UNK-NUM million, or \$1.94 a share.} 
& $0.50$ \\ \cmidrule{2-4}
& The company said it expects to report net income of \$UNK-NUM million, or \$1.04 a share, from \$UNK-NUM  million, or,
& \textit{Nine-month net climbed 19\% to \$UNK-NUM million, or \$2.21 a primary share, from \$UNK-NUM million, or \$1.94 a share.} 
& $0.50$ \\ \cmidrule{2-4}
& The company said it expects to report net income of \$UNK-NUM million, or \$1.04 a share, from \$UNK-NUM  million, or,
& \textit{Nine-month net climbed 19\% to \$UNK-NUM million, or \$2.21 a primary share, from \$UNK-NUM million, or \$1.94 a share.} 
& $0.50$ \\ \midrule
\multirow{3}{*}{+ MDR training} & They have been growing wary of institutional investors. 
 & \textit{People have been very respectful of each other.}
 & $0.46$ \\ \cmidrule{2-4}
& The Palo Alto retailer adds that it expects to post a third-quarter loss of about \$1.8 million, or 68 cents a share, compared
 & \textit{Not counting the extraordinary charge, the company said it would have had a net loss of \$3.1 million, or seven cents a share.} 
 & $0.62$ \\ \cmidrule{2-4}
& But Mr. Chan didn't expect to be the first time in a series of cases of rape and incest, including a claim of two, 
 & \textit{For the year, electronics emerged as Rockwell's largest sector in terms of sales and earnings.}                                    
 & $0.80$ \\ \midrule
\multirow{3}{*}{+ Vamp prior} & But despite the fact that they're losing.  
& \textit{As for the women, they're UNK-LC.}
& $0.45$ \\ \cmidrule{2-4}
& Other companies are also trying to protect their holdings from smaller companies. 
& \textit{And ship lines carrying containers are also trying to raise their rates.}  
& $0.60$ \\ \cmidrule{2-4}
& Dr. Novello said he has been able to unveil a new proposal for Warner Communications Inc., which has been trying to participate in the U.S.
& \textit{President Bush says he will name Donald E. UNK-INITC to the new Treasury post of inspector general, which has responsibilities for the IRS...}
& $0.78$ \\ \midrule
\multirow{3}{*}{+ MoG Prior} & At American Stock Exchange composite trading Friday, Bear Stearns closed at \$25.25 an ounce, down 75 cents. 
& \textit{In American Stock Exchange composite trading yesterday , Westamerica closed at \$22.25 a share, down 75 cents.}
& $0.32$ \\ \cmidrule{2-4}
& Mr. Patel, yes, says the music was ``extremely effective.''
& \textit{Mr. Giuliani's campaign chairman, Peter Powers, says the Dinkins ad is ``deceptive.''}
& $0.57$ \\ \cmidrule{2-4}
& The pilots will be able to sell the entire insurance contract on Nov. 13.  
& \textit{The proposed acquisition will be subject to approval by the Interstate Commerce Commission, Soo Line said.}
& $0.60$ \\ 
\bottomrule
\end{tabular}
\caption{\label{fig:app-samples}Sentences sampled from various models considered in this paper. For the \textsc{RnnLM}, we ancestral-sample directly from the softmax layer. For \textsc{SenVAE}, we sample from the prior and decode greedily. The vanilla \textsc{SenVAE} consistently produces the same sample in this setting, that is because it makes no use of the latent space and all source of variability is encoded in the dynamics of its strong generator. Other \textsc{SenVAE} models were trained with MDR targeting a rate of $10$. Next to each sample we show in \emph{italics} the closest training instance in terms of an edit distance (i.e. TER). The higher this distance (it varies from $0$ to $1$), the more novel the sentence is. This gives us an idea of whether the model is generating novel outputs or copying from the training data.}
\end{figure*}

\begin{figure*}[]
\centering
\small
\fontdimen2\font=1pt
\begin{subfigure}[t]{\textwidth}
\begin{tabular}{@{}p{\textwidth}@{}}
\toprule
\textbf{Revenue rose 12\% to \$UNK-NUM billion from \$UNK-NUM billion.} \\
It is no way to get a lot of ways to get away from its books. \\
At one point after Congress sent Congress to ask the Senate Democrats to extend the bill. \\
So far. \\
But the number of people who want to predict that they can be used to keep their own portfolios, \\
The U.S. government has been announced in 1986, but it was introduced in December 1986 \\
The company said it plans to sell its C\$400 million million shares outstanding \\
\textbf{Revenue slipped 4.6\% to \$UNK-NUM million from \$UNK-NUM million.}\\
\bottomrule
\end{tabular}
\caption{Vanilla \textsc{SenVAE} with ancestral sampling.}
\end{subfigure}
~
\begin{subfigure}[t]{\textwidth}
\begin{tabular}{@{}p{\textwidth}@{}}
\toprule
\textbf{Mr. Vinson estimates the industry's total revenues approach \$200 million.} \\
The company said it expects to report net income of \$UNK-NUM million, or \$1.04 a share, \\
The company said it expects to report net income of \$UNK-NUM million, or \$1.04 a share, \\
The company said it expects to report net income of \$UNK-NUM million, or \$1.04 a share, \\
The company said it expects to report net income of \$UNK-NUM million, or \$1.04 a share, \\
The company said it expects to report net income of \$UNK-NUM million, or \$1.04 a share, \\
The company said it expects to report net income of \$UNK-NUM million, or \$1.04 a share, \\
\textbf{``That's not what our fathers had in mind.''}\\
\bottomrule
\end{tabular}
\caption{Vanilla \textsc{SenVAE} with greedy decoding.}
\end{subfigure}
~
\begin{subfigure}[t]{\textwidth}
\begin{tabular}{@{}p{\textwidth}@{}}
\toprule
\textbf{He could grasp an issue with the blink of an eye.''} \\
He could be called for a few months before the Senate Judiciary Committee Committee. \\
He would be able to accept a clue as the president's argument.  \\
But there is no longer reason to see whether the Soviet Union is interested. \\
But it doesn't mean any formal comment on the basis.  \\
However, there is no longer reason for the Hart-Scott-Rodino Act.  \\
However, Genentech isn't predicting any significant slowdown in the future. \\
\textbf{However, StatesWest isn't abandoning its pursuit of the much-larger Mesa.}\\
\bottomrule
\end{tabular}
\caption{\textsc{SenVAE} trained with \textsc{MDR} ($r=10$). }
\end{subfigure}
~
\begin{subfigure}[t]{\textwidth}
\begin{tabular}{@{}p{\textwidth}@{}}
\toprule
\textbf{Sony was down 130 to UNK-NUM.}\\
 The price was down from \$UNK-NUM.\\
 The price was down from \$ UNK-NUM a barrel to \$UNK-NUM.\\
 The price was down about \$ 130 million.\\
 The yield on six-month CDs rose to 7.93\% from 8.61\%.\\
 Friday's sell-off was down about 60\% from a year ago.\\
 Friday's Market Activity\\
\textbf{Friday's edition misstated the date}\\
 \bottomrule
\end{tabular}
\caption{\textsc{SenVAE} with \textsc{MoG} prior trained with \textsc{MDR} ($r=10$).}
\end{subfigure}
~
\begin{subfigure}[t]{\textwidth}
\begin{tabular}{@{}p{\textwidth}@{}}
\toprule
\textbf{Lawyers for the Garcias said they plan to appeal.} \\
Lawyers for the agency said they can't afford to settle. \\
Lawyers for the rest of the venture won't be reached.   \\
This would be made for the past few weeks. \\
This has been losing the money for their own. \\
This has been a few weeks ago. \\
This has been a very disturbing problem. \\
\textbf{This market has been very badly damaged.''}\\
\bottomrule
\end{tabular} 
\caption{\textsc{SenVAE} with Vamp prior trained with \textsc{MDR} ($r=10$). }
\end{subfigure}
\caption{Latent space homotopies for various \textsc{SenVAE} models. Note the smooth transition of topic and grammatically of the samples in properly trained \textsc{SenVAE} models. Also note the absence of such a smooth transition in the softmax samples from the vanilla \textsc{SenVAE} model.}
\label{fig:app-hom}
\end{figure*}

\end{document}